\pdfoutput=1

\documentclass[11pt]{article}

\usepackage[final]{acl}

\usepackage{times}
\usepackage{latexsym}

\usepackage[T1]{fontenc}

\usepackage[utf8]{inputenc}

\usepackage{microtype}

\usepackage{inconsolata}

\usepackage{amssymb}
\usepackage{subcaption}
\usepackage{caption}
\usepackage{multicol}
\usepackage{multirow}
\usepackage{booktabs}
\usepackage{graphicx}
\usepackage{tabularx}
\usepackage{float}
\usepackage{efbox}
\usepackage{pifont}
\usepackage{xspace,mfirstuc,tabulary}
\usepackage{array}
\usepackage{arydshln}
\usepackage{amsmath}

\DeclareMathOperator*{\argmin}{arg\,min}
\def\@fnsymbol#1{\ensuremath{\ifcase#1\or *\or \dagger\or \ddagger\or
   \mathsection\or \mathparagraph\or \|\or **\or \dagger\dagger
   \or \ddagger\ddagger \else\@ctrerr\fi}}
\newcommand{\ssymbol}[1]{^{\@fnsymbol{#1}}}

\title{Minimizing PLM-Based Few-Shot Intent Detectors}

\author{
Haode Zhang$^1$ \quad \bf Albert Y.S. Lam$^2$ \quad Xiao-Ming Wu$^1$\Thanks{~ Corresponding author.} \\ 
Department of Computing, The Hong Kong Polytechnic University, Hong Kong S.A.R.$^1$ \\
Fano Labs, Hong Kong S.A.R.$^2$ \\
{\tt \small haode.zhang@connect.polyu.hk, xiao-ming.wu@polyu.edu.hk, albert@fano.ai} \\
}

\begin{document}
\maketitle
\begin{abstract}
Recent research has demonstrated the feasibility of training efficient intent detectors based on pre-trained language model~(PLM) with limited labeled data. However, deploying these detectors in resource-constrained environments such as mobile devices poses challenges due to their large sizes. In this work, we aim to address this issue by exploring techniques to minimize the size of PLM-based intent detectors trained with few-shot data. Specifically, we utilize 
large language models (LLMs) for data augmentation, employ a cutting-edge model compression method for knowledge distillation, and devise a vocabulary pruning mechanism called V-Prune. Through these approaches, we successfully achieve a compression ratio of 21 in model memory usage, including both Transformer and the vocabulary, while maintaining almost identical performance levels on four real-world benchmarks.
\end{abstract}

\section{Introduction}

Intent detection refers to the task of classifying users' utterances according to the underlying intents. The task is critical since the result determines the following procedures such as dialogue state tracking, and thus the final output of the dialogue system.
Pre-trained language models (PLMs) make it easier to train an intent detector with only a few labeled data~\cite{zhang2021effectiveness, zhang2022fine}, but these models are difficult to deploy in resource-constrained scenarios such as mobile devices due to the gigantic model sizes.

\begin{figure}[t]
\centering
    \includegraphics[width=0.35\textwidth]{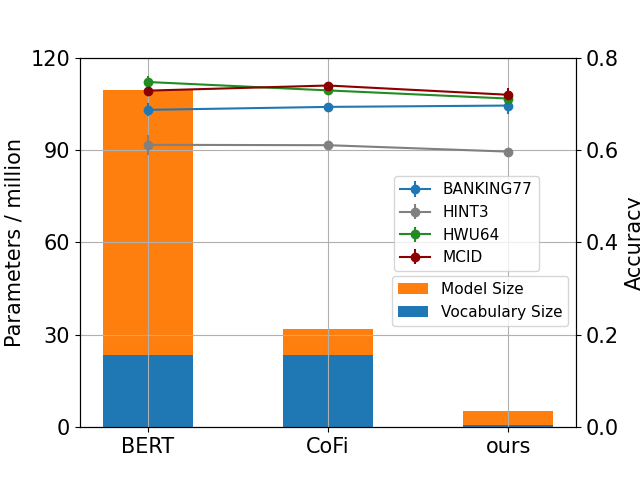}
    \caption{The efficacy of our approach under the 5-shot scenario. Model memory usage is denoted by the stacked bars, while model performance by the lines.}
    \label{figure: main_result_stacked_bar}
\end{figure}

Recent research efforts in PLMs compression mainly consider data-rich scenarios. For instance, ~\citet{xia2022structured} utilize data of up to 393k instances to prune and distill a large encoder into a smaller one. 
\citet{Tao2022CompressionOG} rely on datasets of sizes up to 1.8M to quantize the PLM.
In contrast, PLMs compression under the data-scarce scenario has received limited attention.
\citet{sauer2022knowledge} presumes access to an additional set of annotated intent detection data, which may not be practical. \citet{melas2020generation} fine-tune a generative PLM for data augmentation before model distillation, but the distillation method fails to yield a competitive performance.
Furthermore, current works on PLMs compression almost pay no attention to the vocabulary, despite its significant contribution to the model size. Fig.~\ref{figure: main_result_stacked_bar} illustrates the substantial proportion of the vocabulary within BERT~\cite{devlin2018bert}. \citet{zhao-etal-2021-extremely} and ~\citet{kolesnikovaknowledge} aim to train a task-agnostic model (with the minimum vocabulary size of $5000$), but as to be demonstrated in this study, a model dedicated to intent detection can have a significantly smaller vocabulary.

\begin{figure*}[t]
\centering
    \centering
    \includegraphics[scale=0.47]{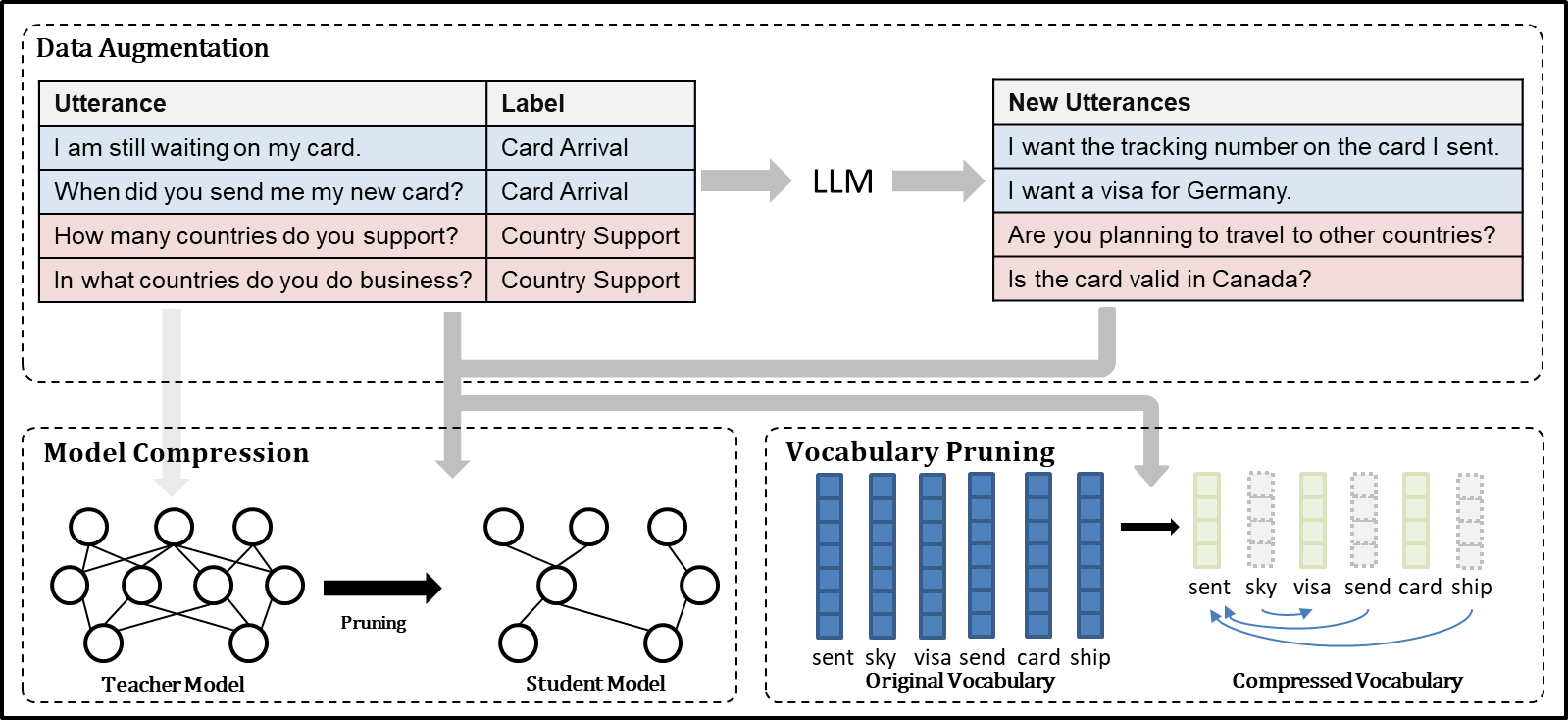}
\caption{Illustration of our method. Off-the-shelf generative language models are adopted to generate new utterances according to the few labeled data. These new data are combined with the few data to compress a large teacher model into a small student model, and also to extract a small vocabulary.}
\label{figure: idea_illustration}
\end{figure*}
To minimize PLM-based few-shot intent detector, we harness the following technology advancements. Specifically, we utilize advanced off-the-shelf generative large language models (LLMs) for data augmentation. Without any fine-tuning~\cite{sauer2022knowledge}, these LLMs are capable of generating high-quality utterances that align with the specified intent.
Subsequently, we employ CoFi, a state-of-the-art Transformer compression method to compress the model. Unlike in \citet{melas2020generation} where the student model is trained from scratch, CoFi gradually prunes the teacher model parameters to derive the student model, and thus yields superior performance.
Finally, we devise a novel vocabulary pruning method, namely V-Prune. V-Prune selectively retains task-relevant tokens in the original vocabulary, and complements the performance drop due to missing tokens during inference via nearest neighbor replacement.
Comprehensive evaluations demonstrate the efficacy of the proposed method. For instance, under the 5-shot scenario, our method reduces the size of the BERT model and its vocabulary size by a factor of 20 and 30 respectively, while almost no loss in the performance is observed over various benchmarks, as shown in Fig.~\ref{figure: main_result_stacked_bar}. The code has been released at~\url{https://github.com/hdzhang-code/smallID/}.

\section{Method}
To train a small intent detector with a small labeled set $\mathcal{D}$, we first train a large teacher model with $\mathcal{D}$ and then distill the knowledge into a small model.

\subsection{Model Compression with Data Augmentation}
\begin{figure}[ht]
    \centering
    \includegraphics[scale=0.45]{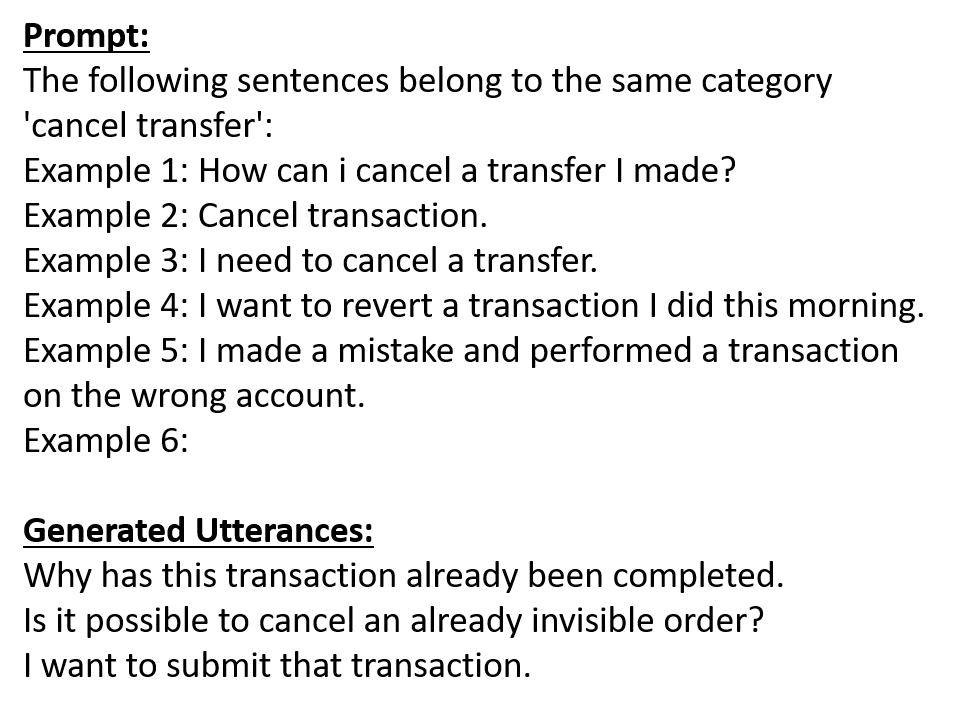}
    \caption{An example of the prompt and generated utterances under the $5$-shot scenario.}
    \label{figure: latex_figures_pilot_study_prompt_example}
\end{figure}

\textbf{Data Augmentation.} To alleviate data scarcity, we adopt data augmentation by off-the-shelf LLMs for two reasons. First, such LLMs do not need to be fine-tuned, which requires additional engineering effort and computation resources. Second, recently published LLMs such as GPT-3 have shown promising results to generate texts with high quality. To prompt LLMs to generate the desired utterances with the specific intent, we adopt the prompt in Fig.~\ref{figure: latex_figures_pilot_study_prompt_example}, wherein some examples of the generated utterances are also given. The generated utterances may have label shifts, but such shifts are tolerated by the model compression method we employ.
\label{Section_Knowledge_Distillation_Data_Augmentation}

\textbf{Model Compression}. To train the teacher model $\text{f}_\text{t}$, we follow the common practice to attach a linear classifier on top of the $\text{[CLS]}$ representation of PLMs~\cite{devlin2018bert, zhang2023revisit} and optimize the parameters $\theta_\text{t}$ with $\mathcal{D}$ and cross-entropy loss function. Then, knowledge distillation is performed via aligning the logit output:
\begin{equation}
  \theta= \argmin_{\theta} \text{KL}\left(\frac{\text{f}\left(\mathcal{D}; \theta\right)}{T}, \frac{\text{f}_\text{t}\left(\mathcal{D}; \theta_\text{t}\right)}{T}\right),
\label{equation: self-distillation KL}
\end{equation}
where $\text{KL}(\cdot)$ is the Kullback-Leibler (KL) divergence, $\text{f}(\cdot)$ and $\text{f}_\text{t}(\cdot)$ denote the output logit of the desired model and the teacher model, respectively. $T$ is the tunable temperature parameter. 
\begin{table*}[t]
\centering
\small
\begin{tabular}{lcccccccc}
\toprule
\multicolumn{1}{l}{\multirow{2}{*}{Method}} & 
\multicolumn{2}{c}{BANKING77} &
\multicolumn{2}{c}{HINT3} &
\multicolumn{2}{c}{HWU64} &
\multicolumn{2}{c}{MCID}  
\\
\cmidrule(lr){2-3}  \cmidrule(lr){4-5}  \cmidrule(lr){6-7} \cmidrule(lr){8-9}
& $90\%$ & $95\%$ & $90\%$ & $95\%$ & $90\%$ & $95\%$ & $90\%$ & $95\%$\\
\midrule
BERT & 68.69\tiny(1.39) & 68.69\tiny(1.39) & 61.12\tiny(2.14)  & 61.12\tiny(2.14) & 74.72\tiny(1.40)  & 74.72\tiny(1.40)  & 72.85\tiny(1.24)  & 72.85\tiny(1.24) \\
\midrule
CNN & 56.44\tiny(.90) & 55.90\tiny(1.30) & 51.21\tiny(1.55)  & 50.68\tiny(1.12) & 59.83\tiny(1.31)  & 59.69\tiny(1.14)  & 55.40\tiny(2.04)  & 54.17\tiny(2.21)  \\
+ GPTJ-6B  & 63.52\tiny(1.20) & 63.37\tiny(1.16) & 58.58\tiny(2.13)  & 59.62\tiny(2.45) & 70.26\tiny(.20)  & 70.35\tiny(.60)  & 69.40\tiny(.50)  & 69.94\tiny(1.22)  \\
+ OPT-30B  & 64.38\tiny(1.82)  & 62.71\tiny(1.44) & 58.04\tiny(.79)  & 58.14\tiny(.67)  & 69.41\tiny(1.21)  & 69.24\tiny(.74) & 70.57\tiny(.59)  & 69.61\tiny(1.15) \\
+ GPT4  & 63.37\tiny(1.69) & 63.79\tiny(1.58) & 56.95\tiny(1.04)  & 56.98\tiny(.66) & 69.68\tiny(1.93)  & 69.15\tiny(.52)  & 67.52\tiny(1.41)  & 69.15\tiny(.52)  \\
\midrule
BiLSTM & 57.75\tiny(1.68) & 59.07\tiny(1.28) & 50.98\tiny(1.54)  & 51.24\tiny(1.44) & 62.17\tiny(1.28)  & 61.94\tiny(1.87)  & 60.53\tiny(2.90)  & 59.10\tiny(2.60)  \\
+ GPTJ-6B  & 68.35\tiny(1.78) & 68.48\tiny(1.86) & 58.43\tiny(1.52)  & 58.61\tiny(.66) & 72.51\tiny(.85)  & 71.66\tiny(.82)  & 67.76\tiny(1.98)  & 68.30\tiny(1.86)  \\
+ OPT-30B  & 69.21\tiny(2.07) & 68.81\tiny(2.14) & 58.79\tiny(1.49)  & 58.20\tiny(.93) & 71.53\tiny(.53)  & 70.91\tiny(.62)  & 67.68\tiny(2.43)  & 67.89\tiny(2.22)  \\
+ GPT4  & 67.39\tiny(1.64) & 66.95\tiny(1.73) & 55.30\tiny(1.58)  & 55.74\tiny(1.31) & 70.54\tiny(1.20)  & 70.17\tiny(.73)  & 64.07\tiny(1.78)  & 64.64\tiny(1.03)  \\
\midrule
CoFi & 69.33\tiny(.02) & 67.05\tiny(.02) & \textbf{61.04\tiny(.02)}  & 59.20\tiny(.03) & 72.90\tiny(.07)  & 69.78\tiny(.01)  & \textbf{73.96\tiny(.02)}  & 67.05\tiny(.02)  \\
+ GPTJ-6B & \textbf{70.67\tiny(1.90)} & \textbf{70.38\tiny(1.80)} & \textbf{62.10\tiny(1.46)}  & \textbf{61.92\tiny(1.88)} & \textbf{74.59\tiny(1.55)}  & \textbf{73.78\tiny(1.14)}  & \textbf{73.80\tiny(1.54)}  & \textbf{72.65\tiny(2.25)}  \\
+ OPT-30B & 70.44\tiny(1.97) & 70.13\tiny(1.79) & \textbf{61.42\tiny(1.35)}  & \textbf{60.83\tiny(1.12)} & 73.88\tiny(1.81)  & 73.38\tiny(1.41)  & 72.36\tiny(2.12)  & 71.29\tiny(1.37)  \\
+ GPT4 & \textbf{71.34\tiny(1.34)} & \textbf{70.86\tiny(1.67)} & 60.89\tiny(1.71)  & \textbf{60.74\tiny(1.32)} & \textbf{74.28\tiny(.53)}  & \textbf{73.76\tiny(.63)}  & 72.07\tiny(2.66)  & \textbf{72.28\tiny(1.78)}  \\
\bottomrule
\end{tabular}
\caption{Evaluation of the proposed framework with different student model architectures. The number of labeled data per label is $5$. Top 3 results are highlighted. + denotes using the following model for synthetic data generation. \%90 and \%95 refer to the compression ratio of the Transformer, excluding the vocabulary.}
\label{table: main_result_distill}
\end{table*}
Vanilla knowledge distillation trains the student model from scratch, and hence results in sub-optimal performance. In this work, we utilize CoFi~\cite{xia2022structured}, a recently proposed Transformer compression method. CoFi gradually prunes both coarse-grained modules and fine-grained parameters of BERT~\cite{devlin2018bert} to obtain the student model, and thus reach a promising performance~\footnote{We refer the reader to the original paper for details.}. 
Like classic knowledge distillation, CoFi adopts the dataset $\mathcal{D}$ on which the output of the small model is aligned with the teacher model. However, when $\mathcal{D}$ is small, a significant performance drop is observed, as to be shown by the experiment.
\subsection{Vocabulary Pruning (V-Prune)}

It is intuitive that the original vocabulary with tens of thousands of tokens is unnecessarily large for an intent detection task, since the task deals with tens of intents, usually under certain scenarios. However, to obtain the task-specific vocabulary, two challenges persist. First, how to estimate the target vocabulary given tens of words in the few labeled utterances. Second, how to complement the performance drop due token missing. To tackle the first issue, we extract the most frequent $K$ tokens in the \emph{augmented dataset} generated in Section~\ref{Section_Knowledge_Distillation_Data_Augmentation}, to compose a vocabulary $V^\prime$. $V^\prime$ is a small fraction of the original vocabulary $V$. For the second challenge, we map the missing token to the nearest tokens in $V^\prime$:
\begin{equation}
        \text{M}(t) = \argmin_{w \in V^\prime} \text{d}(t, w), \forall t \in V, \
    \label{equation: map big vocab to small voab}
\end{equation}
where $\text{M}(t)$ denotes the map from any token $t$ in  $V$, to a token in $V^\prime$. $\text{d}(\cdot, \cdot)$ is the distance function, measuring the semantic distance between two tokens. We use Euclidean distance in our experiment. Fig.~\ref{figure: idea_illustration} gives an example of such mapping.

In addition, to further compress the vocabulary memory footprint, we adopt principal component analysis (PCA) transformation to reduce the dimension of word embeddings. The transformation is applied to map all word embeddings to the low-dimensional space, e.g. from 768 dimensions to 400 dimensions. During inference, the low-dimensional representation is mapped back to the original one by a simple linear mapping, before being fed into the model.

In the following experiment, the vocabulary adopts byte pair encoding (BPE).

\section{Experiments}
\subsection{Experimental Setup}
\paragraph{Datasets.} We adopt $4$ large-scale practical benchmarks, including BANKING77~\cite{casanueva2020efficient}, HINT3~\cite{arora2020hint3}, HWU64~\cite{DBLP:conf/iwsds/LiuESR19} and MCID~\cite{arora2020cross}. We randomly sample $5$ data per label from the training partition to compose $\mathcal{D}$.
\begin{figure*}[t]
\centering
    \begin{subfigure}[t]{0.32\textwidth}
        \includegraphics[scale=0.27]{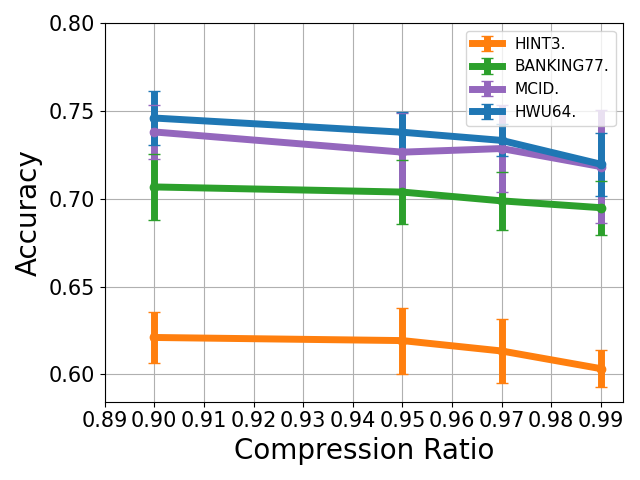}
        \caption{Model size.}
        \label{figure: analysis, Model size}
    \end{subfigure}
    \begin{subfigure}[t]{0.32\textwidth}
        \includegraphics[scale=0.274]{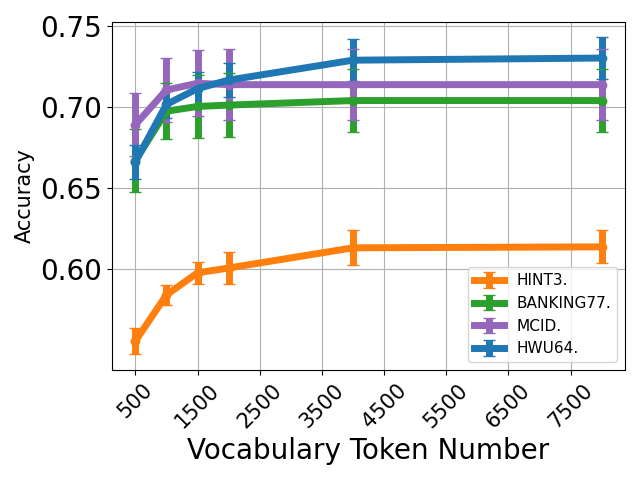}
        \caption{Vocabulary token number.}
        \label{figure: analysis, vocabulary token number}
    \end{subfigure}
    \begin{subfigure}[t]{0.32\textwidth}
        \includegraphics[scale=0.27]{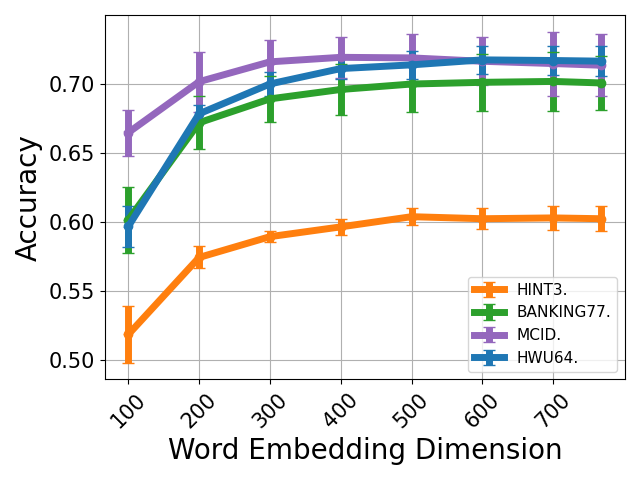}
        \caption{Embedding dimension.}
        \label{figure: analysis, word embedding dimension}
    \end{subfigure}     
\caption{Impact of hyper-parameters on the performance.}
\label{figure: analysis}
\end{figure*}
\begin{table}[t]
\centering
\small
\begin{tabular}{p{0.80cm}>{\centering\arraybackslash}p{1.7cm}p{1.0cm}p{1.0cm}p{0.9cm}}
\toprule
\multicolumn{1}{l}{\multirow{1}{*}{Size$\ssymbol{3}$}} & 
\multicolumn{1}{c}{BANKING77} &
\multicolumn{1}{c}{HINT3} &
\multicolumn{1}{c}{HWU64} &
\multicolumn{1}{c}{MCID}  
\\
\midrule
100\%$\ssymbol{1}$ & 68.69\tiny(1.39) & 61.12\tiny(2.14) & 74.72\tiny(1.40)  & 72.85\tiny(1.24) \\
99.7\%$\ssymbol{5}$ & 70.38\tiny(1.80) & 61.92\tiny(1.88) & 73.78\tiny(1.14)  & 72.65\tiny(2.25) \\
\midrule
3.4\% & 69.62\tiny(1.87) & 59.65\tiny(.59) & 71.13\tiny(.74)  & 71.95\tiny(1.50) \\
\bottomrule
\end{tabular}
\caption{Effectiveness of V-Prune. V-Prune is configured to keep $3.4\%$ vocabulary parameters. $\ssymbol{3}$ denotes the vocabulary size. $\ssymbol{1}$ denotes the vocabulary size of the original BERT. $\ssymbol{5}$ denotes CoFi+GPTJ-6B with the compression ratio of $95\%$.}
\label{table: main_result_vocab}
\end{table}
\paragraph{Methods.} We use the following LLMs to generate data. 
\textbf{GPTJ-6B}~\cite{gpt-j} with 6 billion parameters. 
\textbf{OPT-30B}~\cite{zhang2022opt} with 30 billion parameters. 
\textbf{GPT4}~\cite{openai2023gpt}, one of the most popular LLMs~\footnote{We use the GPT4 variation of text-davinci-003.}.
We also experiment with another two student model architectures: 
\textbf{convolutional neural networks (CNN)}~\cite{chen2015convolutional} and 
\textbf{BiLSTM}~\cite{graves2005framewise}.

\subsection{Results}
\paragraph{Efficacy of model compression.} As shown by Table~\ref{table: main_result_distill}, CoFi with data augmentation achieves the best results, thus demonstrating the performance superiority of our technical choice. Notably as the compression ratio increases to $95\%$, data augmentation enables it to significantly outperform the original CoFi. Furthermore, it surpasses BiLSTM and CNN with the same set of augmented data ~\cite{melas2020generation}, thus justifying our technical choice of CoFi.
Moreover, we have several interesting findings. 
First, the generated data plays a key role in few-shot model compression regardless of the student model architecture. It is noteworthy that when coupled with data augmentation, the simple distillation BiLSTM+GPTJ-6B play on par with CoFi, a highly sophisticated method.
Second, the generative model size does not make a discernible difference for model compression, although it is a popular belief that a larger model generates better text. For example, GPT-J with only 6 billion parameters yields a similar performance to GPT-4 with 30 thousand times larger size.


\paragraph{Efficacy of V-Prune.} We apply V-Prune to CoFi+GPTJ-6B with the compression ratio of $95\%$.
As shown in Table~\ref{table: main_result_vocab}, a tiny fraction of the vocabulary is enough to obtain a decent performance, following the intuition that task-wise vocabulary is small. It is observed that CoFi+GPTJ-6B compresses the vocabulary by a tiny fraction because of the reduction in the word embedding dimension, but V-Prune achieves the compression ratio of 3.4\%.
The reduction in vocabulary size is crucial as the vocabulary constitutes the largest portion of the memory footprint after effective model compression, as shown in Fig.~\ref{figure: main_result_stacked_bar}. 
Additionally, we provide the ablation study result in Table~\ref{table: ablation study, vocabulary pruning}, showing the efficacy of data augmentation and the nearest-neighbor replacement mechanism, respectively.

\begin{table}[h]
\centering
\small
\begin{tabular}{cccccc}
\toprule
5-shot & DA & NN & BANKING77 & HINT3\\
\midrule
\checkmark &  &  & 67.79\tiny(1.80) & 54.44\tiny(1.69)  \\
\checkmark & \checkmark &  & 69.79\tiny(1.62) & 58.85\tiny(1.17) \\
\checkmark & \checkmark & \checkmark & \textbf{70.12\tiny(1.96)} & \textbf{60.09\tiny(.97)} \\
\midrule
5-shot & DA & NN & HWU64 & MCID\\
\midrule
\checkmark &  &  & 67.03\tiny(.73) & 61.15\tiny(2.99)  \\
\checkmark & \checkmark &  & 71.15\tiny(.62) & 69.20\tiny(2.58) \\
\checkmark & \checkmark & \checkmark & \textbf{71.65\tiny(1.06)} & \textbf{71.38\tiny(2.19)} \\
\bottomrule
\end{tabular}
\caption{Ablation study on V-Prune. 5-shot denotes token collection from the  small dataset $\mathcal{D}$. DA denotes the collection with extra data generated by GPT-J. NN denotes the nearest-neighbor replacement mechanism.}
\label{table: ablation study, vocabulary pruning}
\end{table}
\paragraph{Impact of model size on performance.} We visualize the impact of three hyper-parameters controlling the ultimate model size, including compression ratio (Fig.~\ref{figure: analysis, Model size}), vocabulary token number (Fig.~\ref{figure: analysis, vocabulary token number}) and word embedding dimension (Fig.~\ref{figure: analysis, word embedding dimension}). Remarkably, even when the compression ratio reaches as high as $99\%$, the loss in the accuracy remains under $3$ percentage points. Regarding vocabulary size, when the token number decreases to $2000$ and the dimension number to $400$, the performance starts to drop drastically, confirming the conjecture that a small vocabulary is enough for intent detection tasks. The result is notably better than current works on vocabulary reduction, which achieve a minimum vocabulary size of $5000$~\cite{zhao-etal-2021-extremely, kolesnikovaknowledge}. Consequently, we obtain a well-performing intent classifier comprising 5.1 million parameters, which is 21 times smaller than the original BERT, making it convenient to deploy in resource-constrained scenarios.

\section{Conclusion}
In this work, we endeavor to minimize PLM-based few-shot intent detectors. To this end, we explore the cutting-edge PLM compression method CoFi, a data augmentation approach based on generative LLMs, as well as a novel vocabulary pruning technique. The efficacy of our method is demonstrated on four benchmarks, showing its ability to reduce the model size by a factor of 21 without compromising the performance.


\section{Limitations}
The scope of this work is limited, covering only intent detection tasks in English. It is plausible that the proposed approach could be extended to other tasks such as news classification, and also to other languages such as Spanish. Additionally, the generative model employed in this research is large, consuming substantial computational resources to generate new utterances. We will explore these issues in future works.


\appendix

\appendix
\section{Appendix}
\label{sec:appendix}
\subsection{Dataset}
\textbf{BANKING77}~\cite{casanueva2020efficient} is a fine-trained dataset collected in the banking domain with 77 intents.
\textbf{HINT3}~\cite{arora2020hint3} is created from live chatbots with 51 intents. 
\textbf{HWU64}~\cite{DBLP:conf/iwsds/LiuESR19} is a large-scale dataset with 64 intents covering multiple domains. 
\textbf{MCID} ~\cite{arora2020cross} is created for ``Covid-19'' chat bot with 16 intents, across 4 languages, and we use the English part only.
For evaluation under the few-shot scenario, we randomly sample $K$ data per label from the training partition of each dataset to compose dataset $\mathcal{D}$. The statistics are summarized in Table~\ref{table: Dataset statistics}.
\begin{table}[h]
\centering
\small
\begin{tabular}{lcccc}
\toprule
Dataset   &  \#Intent & \#Train & \#Dev & \#Test \\
\midrule
BANKING77 &  77    & 10003  & 0 & 3080    \\
HINT3     &  51    & 1579  & 0 & 676        \\  
HWU64     &  64    & 8954  & 1076 & 1076    \\
MCID      &  16    & 1258 & 148 & 339     \\
\bottomrule
\end{tabular}
\caption{Dataset statistics.}
\label{table: Dataset statistics}
\end{table}

\subsection{Experiment Details}
We use $T = 10$ in Eq.~\ref{equation: self-distillation KL}. For V-Prune, we extract the top 2000 tokens and use the PCA transformation dimension 400. All experiments are performed with NVIDIA's A100 hardware and PyTorch framework.

\end{document}